\documentclass{article}
\usepackage{aikyam}						

\usepackage{booktabs}						
\usepackage{multirow}						
\usepackage{amsfonts}						
\usepackage{graphicx}						
\usepackage{duckuments}						
\usepackage{tcolorbox}

\usepackage[numbers,compress,sort]{natbib}	

\title[Do Vision Language Models Need to Process
Image Tokens?]{Do Vision Language Models Need to Process Image Tokens?}

\author[Ghosh et al.]{
Sambit Ghosh\textsuperscript{1,2} \quad
R. Venkatesh Babu\textsuperscript{2} \quad
Chirag Agarwal\textsuperscript{3} \\
\textsuperscript{1}IBM 
\textsuperscript{2}Indian Institute of Science 
\textsuperscript{3}University of Virginia \\
\texttt{\{sambitghosh, venky\}@iisc.ac.in, chiragagarwal@virginia.edu}
}

\newcommand{\xhdr}[1]{\vspace{0em}\noindent{{\bf #1.}}}
\newcommand{\ie}{\textit{i.e., \xspace}}
\newcommand{\eg}{\textit{e.g., \xspace}}

\newcommand{\hide}[1]{}

\usepackage{colortbl}

\definecolor{Gray}{gray}{0.9}
\definecolor{LightCyan}{rgb}{0.88,1,1}
\definecolor{darkred}{rgb}{0.8,0.1,0.1}
\definecolor{darkyellow}{rgb}{0.95, 0.68, 0.22}
\definecolor{darkgreen}{rgb}{0.1,0.8,0.1}
\newcolumntype{a}{>{\columncolor{Gray}}c}
\newcolumntype{b}{>{\columncolor{white}}c}

\begin{document}

\maketitle

\begin{figure*}[h]
    \centering
    \includegraphics[width=\textwidth]{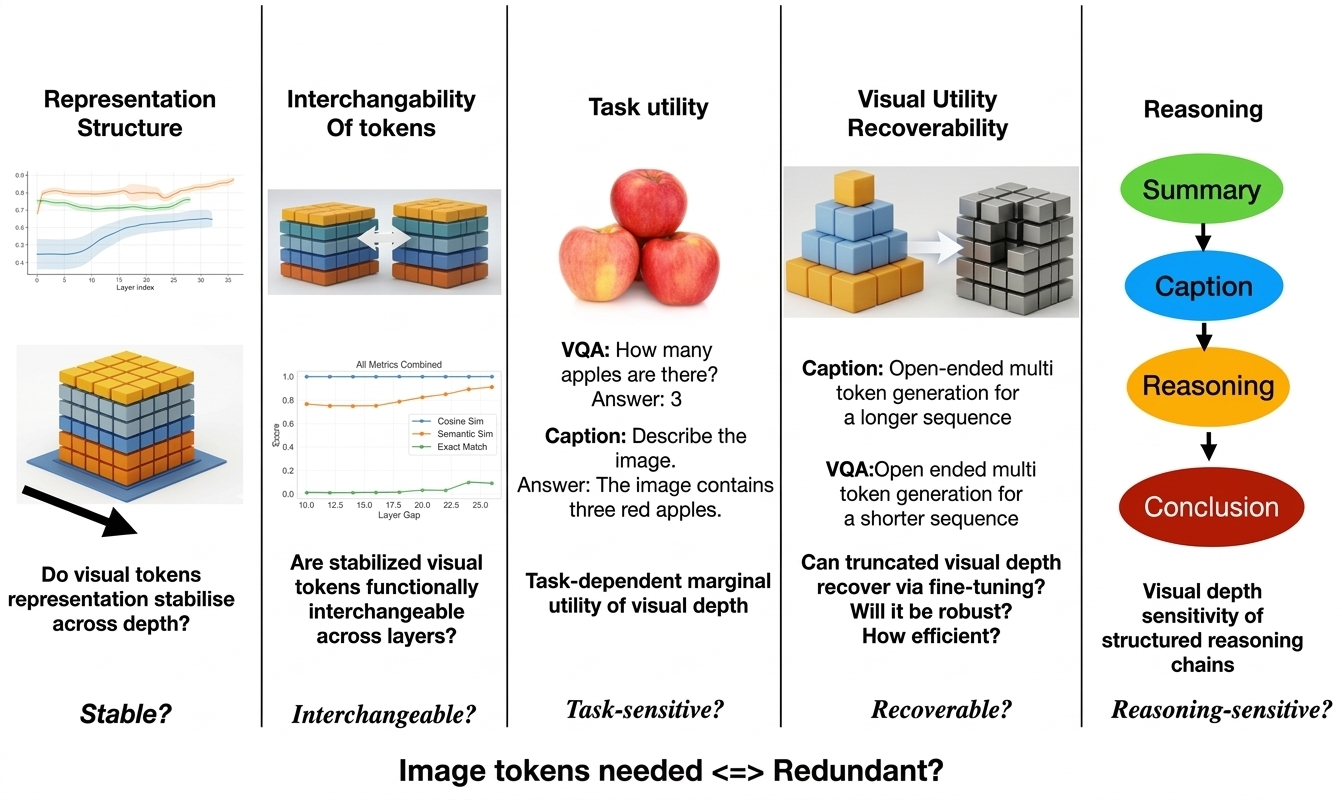}
    \vspace{-0.5cm}
    \caption{\looseness=-1\textbf{When is visual depth necessary in vision language models?} We investigate \textbf{a)} stabilization of image tokens, \textbf{b)} layer-wise functional interchangeability, \textbf{c)} task-dependent utility, \textbf{d)} recoverability after truncation, \textbf{e)} reasoning-chain sensitivity. Our results reveal that image representations stabilize early, but their necessity depends on task and reasoning  demands.
    }
\end{figure*}

\begin{abstract}
    \looseness=-1 Vision Language Models (VLMs) have achieved remarkable success by integrating visual encoders with large language models (LLMs). While VLMs process dense image tokens across deep transformer stacks (incurring substantial computational overhead), it remains fundamentally unclear whether sustained image-token processing is necessary for their performance or visual representations meaningfully evolve from early to later layers. In this work, we systematically investigate the functional role of image tokens in VLMs and show that visual representations rapidly converge to a bounded-complexity regime, \ie their entropy stabilizes, intrinsic dimensionality compresses, and trajectory curvature approaches a near-constant profile. In contrast, textual representations continue to undergo substantial restructuring across depth. Once stabilized, visual representations become largely interchangeable between layers, indicating limited additional transformation in deeper stages. Further, depth-wise visual truncation reveals that the necessity of visual processing is task-dependent, where single-token predictions remain comparatively robust to truncated visual depth, but multi-token generation require sustained access to visual representations. Under deterministic decoding, reducing visual depth perturbs intermediate reasoning trajectories more strongly than final outputs, suggesting that image tokens influence the structure of reasoning more than the ultimate conclusions. Collectively, these findings \textbf{question the assumption} that deeper visual processing is uniformly essential in VLMs, challenging the current paradigm of multimodal LLM architectures. The code can be found on \href{https://github.com/sambitghsh/VLM-Token-Process}{here}.    
\end{abstract}

\section{Introduction}
\label{sec:intro}

Vision–language models (VLMs) \cite{liu2024improved, bai2025qwen25vltechnicalreport, zhu2025internvl3} combine large language models (LLMs) with powerful vision encoders to enable multimodal reasoning over images and text. In a standard VLM pipeline, an input image is divided into fixed-size patches and processed by a visual encoder~\cite{radford2021learning} to produce a sequence of image tokens. These visual representations are then projected into the language model's embedding space through a learned projection layer, allowing the transformer to jointly process image and textual tokens.

Despite incorporating both modalities, recent evidence suggests that visual signals may be \underline{underutilized} during multimodal tasks~\cite{fu2025hidden, deng2025words, rahmanzadehgervi2024vision, hegde2023analyzing,seth2025hallucinogen,seth2025egoillusion,agarwal2025rethinking}, with textual tokens often dominating the final predictions. This raises a fundamental question: \textbf{how does visual information evolve inside VLMs, and when does it remain necessary?}

In this work, we take a representation-centric perspective and analyze how image and textual token representations evolve across the depth of VLM decoders. Rather than evaluating only task-level performance, we investigate the \textbf{structural organisation of token representations} to understand when image tokens contribute meaningful information and when they become redundant. Specifically, we investigate the following research questions:
\begin{enumerate}
    \item \xhdr{Representational properties} How do image and textual token representations evolve across transformer layers?
    \item \xhdr{Interchangeability} Does early stabilization of visual representations imply visual functional redundancy in deeper layers?
    \item \xhdr{Task dependence} Is the necessity of image tokens dependent on the output task, particularly when comparing single-token prediction with multi-token generation tasks such as captioning?
    \item \xhdr{Adaptation} Can fine-tuning recover performance after removing structurally redundant image tokens? Here, we also investigate the model's computational efficiency.
    \item \xhdr{Reasoning} Can structured reasoning chains compensate for reduced visual-token processing depth?
\end{enumerate}
To systematically analyze multimodal representations, we employ matrix entropy, intrinsic dimensionality, and trajectory curvature~\cite{skean2025layer, johnsson2016structures} to trace the structural evolution of token representations across layers and quantify changes in spectral concentration, redundancy, and manifold complexity.

\looseness=-1 Our work provides a \textbf{representational understanding of image-token dynamics in VLMs}, where we show that image tokens exhibit early representation stabilization, become partially interchangeable across depth, and display strong task-dependent utility. Additionally, our results across diverse VLM families and datasets show that fine-tuning can partially redistribute visual information after truncation and approximate the behaviour of the original model, although recovery remains gradual and limited under aggressive truncation. By grounding image-token reduction in representation analysis, our framework provides new insights into multimodal representation dynamics and offers guidance for designing \textbf{more efficient vision-language models}.

\section{Related Work}
\label{sec:related}
Recent work has explored improving the efficiency of vision–language models by reducing the number of image tokens processed during inference. Modern multimodal architectures such as LLaVA \cite{liu2024improved}, Qwen-VL \cite{bai2025qwen25vltechnicalreport}, and InternVL \cite{zhu2025internvl3} integrate vision encoders with large language models but require processing large numbers of visual tokens, leading to significant computational overhead. To mitigate this cost, several works propose pruning strategies that attempt to remove redundant visual tokens while preserving performance \cite{ye2025fit, deng2025scope, zhao2025pore, zhang2024sparsevlm}. This direction is closely related to earlier studies on efficient vision transformers and token sparsification \cite{rao2021dynamicvit, bolya2022token}, which demonstrate that many tokens contribute limited additional information during deep transformer computation. \cite{shi2025vision} demonstrate only some layers are functional while making inferences.

Another line of work studies redundancy across transformer depth. Approaches such as \cite{yuan2025shortv, suo2025pruning, tong2025flowcut} suggest that visual representations exhibit limited novelty in deeper layers and explore exploiting this redundancy for more efficient inference. Closely related to our setting, \cite{lin2025boosting, chen2024image} propose dynamically removing visual tokens as computation progresses through the transformer.

\looseness=-1 Unlike prior works that primarily assume visual redundancy and design pruning mechanisms accordingly, our work focuses on characterising the representational dynamics underlying such redundancy. Inspired by \cite{skean2025layer}, we systematically analyze how image-token representations evolve across transformer depth, examine when they become functionally interchangeable, and study the extent to which truncated models can recover the behavior of the original model through distillation-based adaptation.
\section{RQ1: How do representations evolve across modalities?}
\label{sec:rq1}

\begin{figure}[t]
    \centering
    \includegraphics[width=\columnwidth, height=0.18\textheight]{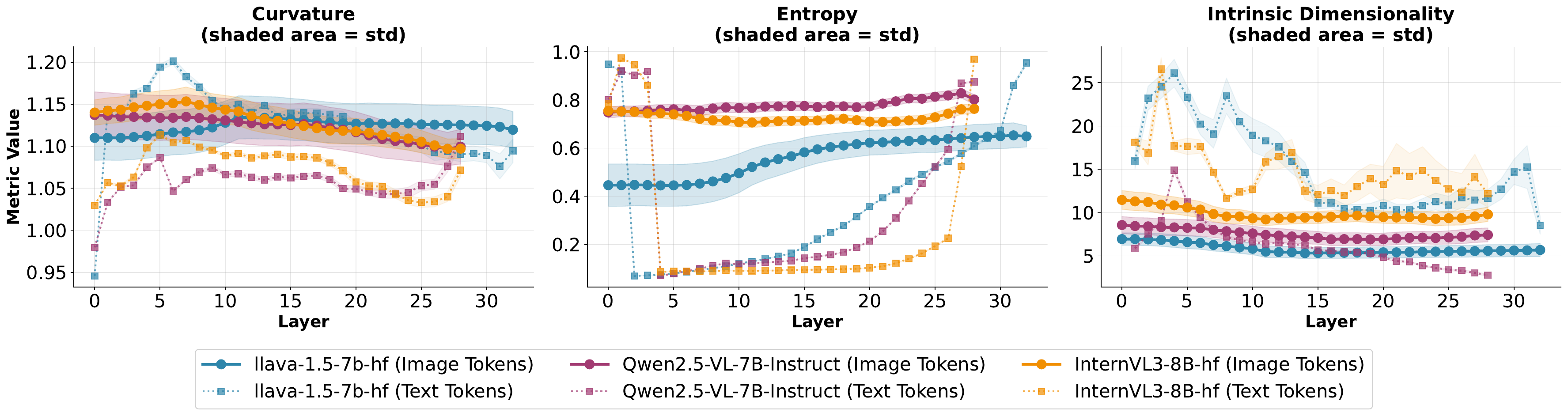}
    \caption{
    Layer-wise evolution of matrix entropy, intrinsic dimensionality, and trajectory curvature 
    for image and textual tokens. Image tokens exhibit early stabilization across all three metrics,
    whereas textual tokens remain dynamically evolving across depth.
    }
    \label{fig:ent_curv_id}
\end{figure}

\looseness=-1 \xhdr{Motivation} Recent works \cite{fu2025hidden, deng2025words, rahmanzadehgervi2024vision} show that vision–language models underperform their underlying vision encoders on pure vision tasks, suggesting that visual information may not be fully utilised in multimodal reasoning. This motivates a closer examination of how visual signals evolve within multimodal architectures and whether deeper processing introduces genuinely new information beyond early representations. We therefore analyze the layer-wise dynamics of image and textual token representations in the VLM decoder, focusing on representational structure rather than task-level accuracy.

\looseness=-1\xhdr{Experimental Setup} We study three VLM families, LLaVA-1.5~\cite{liu2024improved}, Qwen2.5-VL-Instruct \cite{bai2025qwen25vltechnicalreport}, and InternVL~\cite{zhu2025internvl3}, spanning parameter scales from 3B to 72B. Experiments are conducted on the BLINK dataset~\cite{fu2024blink}. To ensure controlled comparison, all qualitative analyzes use identical images across captioning and VQA-style prompts, isolating representational effects from input variation. See Appendix~\ref{app:rq1} for more experiment details. Following~\cite{skean2025layer, johnsson2016structures}, we examine three complementary layer-wise metrics, \ie matrix entropy, intrinsic dimensionality, and trajectory curvature, to characterise how representational capacity is allocated and reorganised across depth, defined as follows:\vspace{0.05in}

\noindent\textit{1) Matrix-Based Entropy and Spectral Structure:} Let $Z_l \in \mathbb{R}^{N \times d}$ denote the layer-$l$ token embeddings. We form the Gram matrix
\begin{equation}
K_l =
\begin{cases}
Z_l^\top Z_l, & N > d, \\
Z_l Z_l^\top, & \text{otherwise},
\end{cases}    
\end{equation}
\looseness=-1 and $\quad\tilde{K}_l = \frac{K_l}{\mathrm{tr}(K_l)}$ \ie the eigenvalues $\{\lambda_i^{(l)}\}$ sum to one. Next, we compute the Von Neumann entropy: $S_1(Z_l) = -\sum_i \lambda_i^{(l)} \log \lambda_i^{(l)},$ and define the effective rank as $r_{\mathrm{eff}}(Z_l) = \exp(S_1(Z_l))$. In this context, the image and text modalities inside the decoder LLM are analyzed separately. Entropy quantifies spectral concentration: \textit{low entropy indicates energy concentrated in a few dominant directions (compressed structure), while high entropy reflects energy distributed across many directions (greater spectral diversity)}.\vspace{0.05in}


\noindent\textit{2) Intrinsic Dimensionality (ID).} It quantifies the effective degrees of freedom of the representation by estimating the dimensionality of the local manifold on which $Z_l$ lies. Unlike ambient dimension, ID captures the minimal number of directions required to describe local variation in the embeddings, thereby reflecting representational complexity independent of the embedding size.\vspace{0.05in}

\noindent\textit{3) Trajectory Curvature (TC).} We quantify curvature as the mean angular deviation between successive displacement vectors:
\begin{equation}
\bar{C}_l =
\frac{1}{N}
\sum_{i=1}^{N}
\arccos
\left(
\frac{\langle v_l^{(i)}, v_{l-1}^{(i)} \rangle}
{\|v_l^{(i)}\| \, \|v_{l-1}^{(i)}\|}
\right),
\end{equation}
where $v_l^{(i)} = z_{l+1}^{(i)} - z_l^{(i)}$ denotes the layer-to-layer update of token $i$. TC captures the degree of directional reconfiguration in embedding space: \textit{high curvature reflects sharp, locally varying updates, whereas low curvature indicates smoother, more globally consistent representational evolution}.\vspace{0.05in}

\looseness=-1 \xhdr{Findings} In Fig.~\ref{fig:ent_curv_id}, across architectures, image tokens consistently enter a stable geometric regime. After a brief adjustment phase, entropy remains confined within a narrow band, effective rank plateaus, intrinsic dimensionality stabilizes, and curvature stays nearly constant. Together, these trends indicate gradual compression followed by convergence to a bounded-complexity manifold, where subsequent layers primarily preserve and refine visual information rather than fundamentally restructuring it. In contrast, textual tokens remain dynamically reorganized throughout depth. Their entropy exhibits sustained fluctuations with repeated spectral redistribution, intrinsic dimensionality alternates between expansion and compression, and curvature is larger and more variable, particularly in intermediate layers. These patterns reflect ongoing semantic transformation rather than early convergence.

Taken together, all three metrics reveal a consistent structural asymmetry: \textit{visual representations stabilize into a bounded structural regime, whereas textual representations continue to evolve and reorganise across layers}.
This pattern is consistent across \textbf{six} models and the convergence of all metrics points to a structural property of multimodal transformers rather than a scale-dependent artefact, raising a natural question: \textit{\textbf{if visual representations stabilize structurally, do deeper layers meaningfully transform them, or do they become functionally interchangeable across depth?}}


\begin{tcolorbox}
    \textbf{Key Takeaway.}  
Image tokens reach a stable representational regime early in the network, whereas textual tokens continue to be dynamically refined across layers.
\end{tcolorbox}
\section{RQ2: Does structural stabilization correspond to layer-wise functional interchangeability?}
\label{sec:rq2}

\begin{figure}[t]
    \centering
    
    \includegraphics[width=0.9\columnwidth]{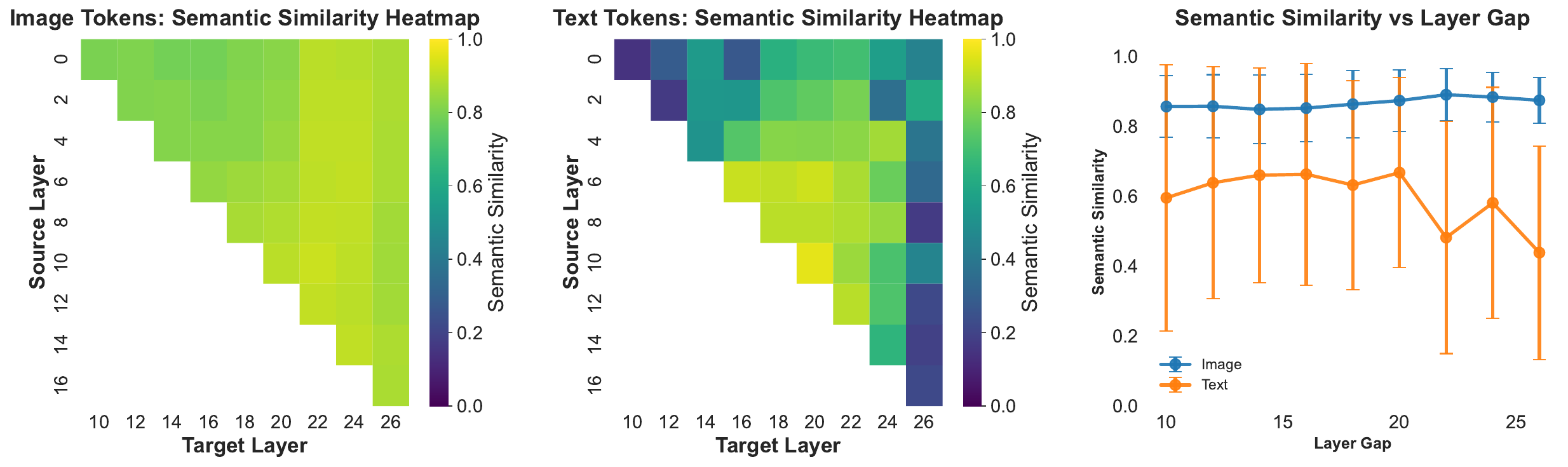}
    
    
    
    
    
    \caption{
  Layer-substitution analysis for captioning with Qwen2.5-VL-7B-Instruct.
For each row, we show the image-token similarity heatmap (left), text-token similarity heatmap (center), and the semantic similarity as a function of the layer gap (right). Additional results are reported in Appendix~\ref{app:rq2}}
    \label{fig:layer_sweep_all}
\end{figure}
\xhdr{Motivation} In this section, we examine whether the geometric stabilization observed in \textbf{RQ1} has functional implications. Building on the convergence of visual representations to a bounded structural regime, we test whether embeddings from different depths are functionally interchangeable, \ie whether substituting visual tokens across layers preserves output semantics. 

\xhdr{Experimental Setup} All experiments in this section use the same image inputs and model families as in \textbf{RQ1} to ensure controlled comparisons. For captioning, we compute semantic similarity between the original and outputs; for VQA, we use Exact Match. A hybrid state is defined as a representation where image tokens from layer $l_a$ and textual tokens from layer $l_b$ are combined and propagated through the remaining decoder layers. If visual representations are truly depth-stable, substitutions from earlier layers should maintain semantics even for large layer gaps $|l_a - l_b|$. In both cases, the unmodified base model output serves as the reference.\vspace{0.05in}

\noindent\textbf{\textit{Layer Substitution Protocol:}} Let $Z_l^{\mathrm{img}}$ and $Z_l^{\mathrm{txt}}$ denote the image and textual token embeddings at layer $l$. To test functional stability, we construct hybrid states by replacing later-layer embeddings with those from earlier layers (which are less processed). For layers $(l_a, l_b)$ with $l_a < l_b$, we form $Z_{\mathrm{hybrid}} = \left( Z_{l_a}^{\mathrm{img}}, \; Z_{l_b}^{\mathrm{txt}} \right)$,
while keeping all other components unchanged, and forward the modified representation through the remaining decoder layers.\vspace{0.05in}

\looseness=-1\xhdr{Findings} In Fig. \ref{fig:layer_sweep_all}, we observe that image-token substitution uniformly maintains high semantic similarity across layer pairs. The similarity heatmaps remain consistently strong across shallow and deep layer combinations. Moreover, the similarity curves as a function of layer gap remain nearly flat ($\approx 1.0$), indicating that visual embeddings extracted from different depths produce \textbf{nearly identical} semantic outputs when substituted, \ie once visual representations stabilize structurally, later layers do not substantially alter their semantic contribution.
In contrast, text-token substitution results in clear degradation as layer gap increases. Off-diagonal regions in the heatmaps exhibit lower similarity values, and the similarity curves decrease steadily with increasing depth difference. Textual representations are therefore \textbf{not interchangeable across depth}. Their semantic role depends critically on the specific layer at which they are extracted, reflecting ongoing transformation and reconfiguration throughout the stack. We observe similar trends across all different model families (see Appendix~\ref{app:rq2} for results on LLaVA-1.5-7B and InternVL3-8B, as well as on different tasks).
Despite architectural differences, image-token substitution is largely depth-invariant, whereas textual substitution remains depth-sensitive. This cross-model consistency suggests a structural property of multimodal transformers rather than a model-specific artefact.

\looseness=-1 The functional interchangeability of image tokens mirrors the structural stabilization observed in \textbf{RQ1}: \textit{once matrix entropy, intrinsic dimensionality, and curvature plateau, deeper layers introduce limited semantic change}. Accordingly, substituting visual representations across depth preserves output semantics, as later layers largely maintain rather than restructure visual information. 
Notably, interchangeability does not imply redundancy, but rather that beyond a certain depth, \textbf{further processing yields minimal semantic modification}.



\begin{tcolorbox}
\textbf{Key Takeaway.}  
   Representational stabilization of image tokens corresponds to layer-wise functional interchangeability. Visual representations exhibit bounded semantic sensitivity to depth substitution, whereas textual representations remain strongly depth-dependent.
\end{tcolorbox}

This observation naturally motivates our next research question: \textit{\textbf{are image tokens still necessary at later layers, or can they be safely truncated without affecting model behavior?}}
\section{RQ3: Task-Dependent Utility of Image Tokens}
\label{sec:rq3_task_dependence}

\begin{figure}[t]
    \centering
    \includegraphics[width=\columnwidth]{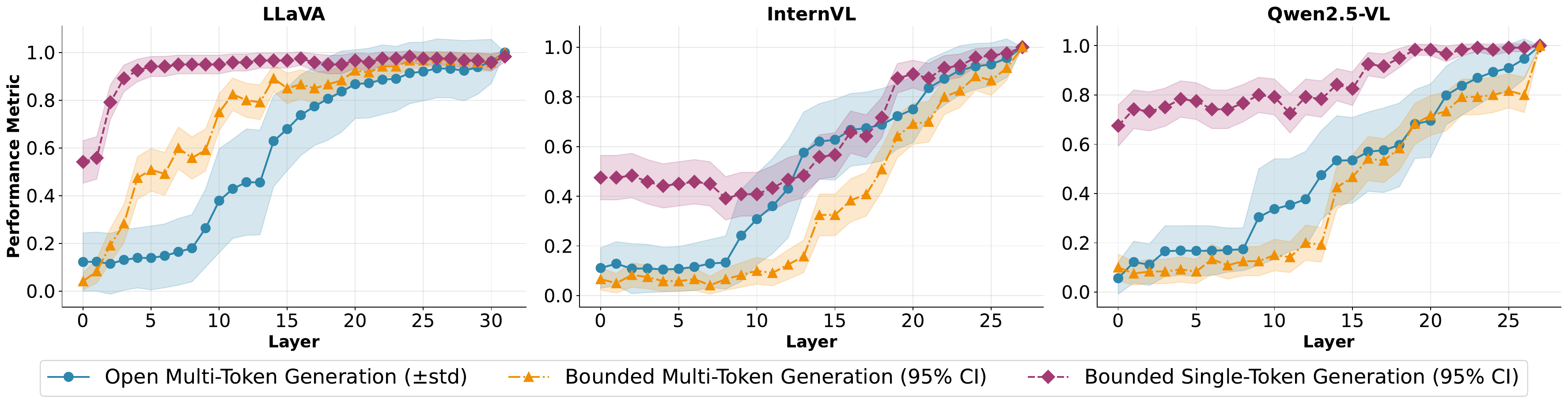}
    \caption{\looseness=-1 Depth-dependent performance under image-token removal. We evaluate three output regimes on the same visual inputs: MCQ (bounded single-token generation), VQA (index+answer match, bounded multi-token generation), and caption generation.
    }
    \label{fig:rq3_pruning}
    \vspace{-0.749cm}
\end{figure}
\xhdr{Motivation} \textbf{RQ1} showed that visual representations undergo geometric stabilization, and \textbf{RQ2} established that, beyond this point, image tokens are largely interchangeable across layers. However, interchangeability alone does not imply that image tokens are dispensable for generation. We therefore investigate how the contribution of image tokens depends on the structure of the output being generated and up to which depth their processing remains functionally necessary.\vspace{0.05in}

\xhdr{Experimental Setup} All experiments use the same image inputs and model families as in \textbf{RQ1} to maintain controlled and consistent comparisons. Similar to RQ1 and RQ2, we evaluate two generation regimes: visual question answering (VQA) and captioning. VQA includes both single-token generation, evaluated with \textit{exact match accuracy}, and open-form responses, measured using \textit{Index + Answer Exact Match}. For captioning (multi-token generation), we report semantic similarity alongside BLEU ~\cite{papineni2002bleu} and ROUGE (ROUGE-1/2/L) scores~\cite{grusky2023rogue}. For more results and details about the metrics see Appendix~\ref{app:metric}. \vspace{0.05in}

\noindent\textbf{\textit{Image-Token Removal Protocol:}} We select a cut layer $l_c$ and terminate image-token processing beyond this depth. Let $H_l \in \mathbb{R}^{B \times N_l \times d}$ denote the hidden states at layer $l$, where $B$ is the batch size, $N_l$ the sequence length at layer $l$, and $d$ the embedding dimension. Let $\mathcal{I}_{\mathrm{img}}$ and $\mathcal{I}_{\mathrm{txt}} \subset \{1,\dots,N_l\}$ denote the index sets of input and textual tokens, identified from the input IDs.

For all $l > l_c$, we restrict the representation to textual tokens by applying, $H_l \leftarrow H_l[:, \mathcal{I}_{\mathrm{txt}}, :]$, thereby removing image-token activations from subsequent computation. By removing all image tokens, we obtain a controlled ablation that isolates visual-depth processing, independent of token-selection heuristics. The same index restriction is consistently applied to position IDs, attention masks, cache positions, and positional embeddings to preserve autoregressive and attention structure under the reduced sequence. The truncated states are then propagated through the remaining decoder layers. By sweeping $l_c$ across depth, we directly quantify how continued image-token processing influences the model’s final predictions.\vspace{0.05in}

\looseness=-1 \xhdr{Findings} Results in Fig.~\ref{fig:rq3_pruning} show that performance declines in both single-token and multi-token settings as image tokens are truncated earlier, but multi-token is markedly more sensitive. Single-token accuracy degrades smoothly and recovers monotonically with retained visual depth, indicating that coarse decision boundaries can be formed with limited visual processing. In contrast, full-string correctness (Index + Answer exact match) exhibits substantially larger degradation under early truncation, highlighting the stronger reliance of fine-grained lexical fidelity on deeper visual representations. 

Further, we observe that captioning consistently exhibits a large sensitivity to early image-token truncation across all models. As shown in Fig.~\ref{fig:rq3_pruning}, semantic similarity drops sharply when image tokens are removed in shallow layers and improves progressively as deeper layers retain visual processing. Performance approaches the full model only when visual tokens are preserved through later stages of the decoder.
This trend is consistent across BLEU~\cite{papineni2002bleu} and ROUGE metrics \cite{grusky2023rogue}(reported in Appendix~\ref{app:rq3}), which show the same monotonic improvement as the $l_c$ moves deeper, where early truncation results in substantial lexical and semantic degradation. 
Across models, LLaVA\cite{liu2024improved} exhibits the strongest recovery as visual depth increases, whereas InternVL\cite{zhu2025internvl3} shows the weakest recovery, indicating greater sensitivity to early visual truncation.

Since all tasks use identical visual inputs, the differing degradation patterns arise from output structure rather than input variation. The depth at which performance stabilizes shifts systematically with output complexity and evaluation strictness. Single-token generation requires only a coarse visually grounded decision; once established, \textbf{further visual processing provides limited benefit}. In contrast, sequence generation involves repeated visual conditioning across decoding steps, increasing sensitivity to early image-token truncation.

\begin{tcolorbox}
\textbf{Key Takeaway.}
\looseness=-1 The required depth of image-token processing scales with output complexity: single-token prediction tolerates early truncation, multi-token generation requires visual retention through deeper layers, and VQA shows high sensitivity depending on evaluation strictness.
\end{tcolorbox}

\section{RQ4: Can a truncated model recover the functional behavior of the original model through fine-tuning?}
\label{sec:rq4}
\xhdr{Motivation} RQ3 showed that truncating image-token processing leads to depth-dependent performance degradation. Next, we investigate whether this degradation is fundamentally irreversible or whether a truncated model can re-approximate the original input-output mapping through adaptation.\vspace{0.05in}

\looseness=-1\xhdr{Experimental Setup} We conduct these experiments on two instruction-tuned models (LLaVA-1.5-7B and Qwen2.5-VL-7B-Instruct) using LoRA fine-tuning~\cite{hu2022lora}. Because truncation effects are most pronounced in open-ended generation, where outputs are not constrained by predefined options, we focus on free-form VQA (ChartQA~\cite{masry2022chartqa} without answer choices) and captioning (Flickr8k). In both tasks, responses must be generated without multiple-choice anchoring, making recovery dependent on preserved visual grounding rather than language priors.\vspace{0.05in}

\looseness=-1\noindent\textbf{\textit{Distillation-Based Fine-Tuning Setup.}} For each image-token truncation depth $K$, let $f_{\text{base}}$ denote the full model and $f_K$ the truncated model. We fine-tune $f_K$ to obtain $\tilde{f}_K$ using the base model's outputs as targets: $y_{\text{target}} = f_{\text{base}}(x)$. The optimisation objective is to minimize the discrepancy between $\tilde{f}_K(x)$ and $f_{\text{base}}(x)$, rather than between model outputs and human annotations. This formulation enables us to examine the \emph{reversibility of representational loss} induced by truncation. If there exist parameters $\theta_K^*$ such that $\tilde{f}_K(x; \theta_K^*) \approx f_{\text{base}}(x)$. Under a task-appropriate metric $\mathcal{M}$, then the information removed by truncation is functionally recoverable. Conversely, persistent misalignment indicates irreversible loss. Alignment is evaluated using Exact Match for VQA and,  semantic and lexical similarity for captioning.\vspace{0.05in}

\begin{figure}[t]
    \centering
    \includegraphics[width=\columnwidth]{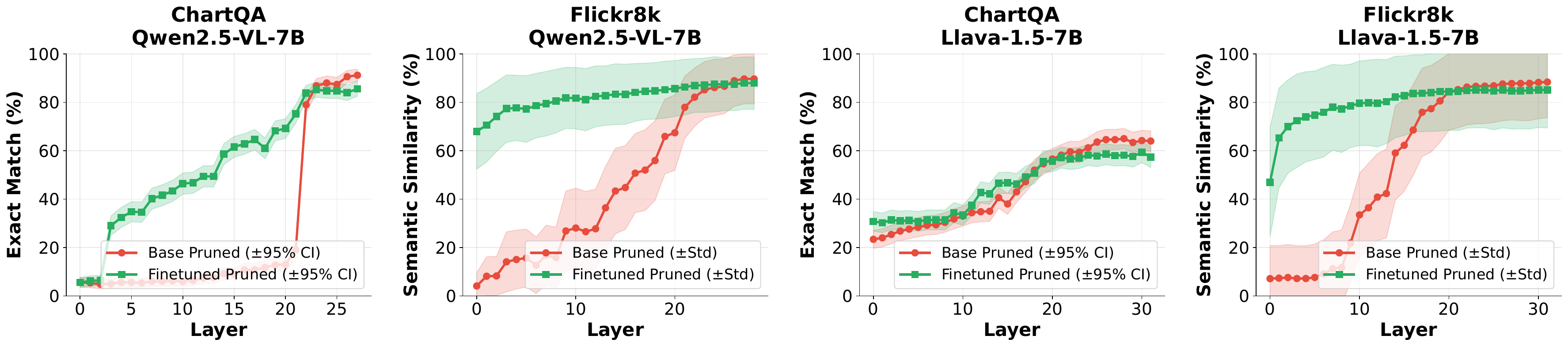}
    \caption{Functional recovery on Flickr8K, ChartQA after distillation-based fine-tuning. Additional details of the result in Appendix-\ref{app:rq4}}
    \label{fig:chartqa_recovery}
\end{figure}

\xhdr{Findings}
In Fig.~\ref{fig:chartqa_recovery}, we observe that, for captioning, fine-tuning restores performance substantially across most truncation depths.
Early truncation limits full restoration, whereas deeper truncation enables near-complete alignment. The recovery curve is continuous rather than segmented into discrete redundancy phases, indicating gradual redistribution of representational capacity rather than sharply bounded critical layers.
For ChartQA without answer options, fine-tuning yields improvements in exact match compared to the truncated models (Fig.~\ref{fig:chartqa_recovery}), but the magnitude of recovery is substantially smaller than in captioning. When image tokens are truncated aggressively at shallow layers, full restoration is not achieved. A likely explanation is that exact match is a discontinuous metric: even minor deviations in numerical or categorical outputs result in zero credit. If early visual-processing layers responsible for precise structured grounding are removed, fine-tuning cannot fully reconstruct the missing fine-grained alignment, limiting recoverability. Let $R(K)$ denote the post-fine-tuning alignment with the base model with $L$ the number of layers at truncation depth $K$. Empirically, $\frac{dR}{dK} > 0,$ indicating that recoverability increases monotonically with retained visual depth. However, the asymptotic behavior is task-dependent:
\[
\lim_{K \to L} R_{\text{caption}}(K) \approx R_{\text{full}}, 
\qquad
\lim_{K \to L} R_{\text{chart}}(K) < R_{\text{full}} \;\; \text{under early truncation}.
\]
thus, although recovery improves smoothly in both settings, captioning achieves substantially higher asymptotic restoration than ChartQA. This suggests that coarse semantic generation is more redistributable under truncation, whereas precise visual-text alignment required for structured question answering imposes stricter irreversibility constraints.\vspace{0.05in}
\noindent\textbf{\textit{Stability Under Decoding Variation:}} We next evaluate whether fine-tuning improves generation stability under different decoding strategies. Rather than testing invariance across sampling and decoding methods, we measure the sensitivity of model outputs to decoding and sampling variation. Let $\mathcal{D}$ denote decoding strategy (greedy, beam, nucleus, top-$k$, temperature sampling). For a fixed truncation depth $K$, let performance under decoding method $\mathcal{D}$ be: $\mathcal{M}_{\mathcal{D}}(K) = \mathcal{M}(\tilde{f}^{\mathcal{D}, t}_K(x), f_{\text{base}}(x))$. We quantify cross-decoding variability using the coefficient of variation~\cite{abdi2010coefficient}: $\mathrm{CV}(K) = \frac{\sigma_{\mathcal{D}}(K)} {\mu_{\mathcal{D}}(K)}$, where $\mu_{\mathcal{D}}(K)$ and $\sigma_{\mathcal{D}}(K)$ denote the mean and standard deviation across decoding strategies.\vspace{0.03in}

\begin{figure}[t]
    \centering
    \includegraphics[width=\columnwidth]{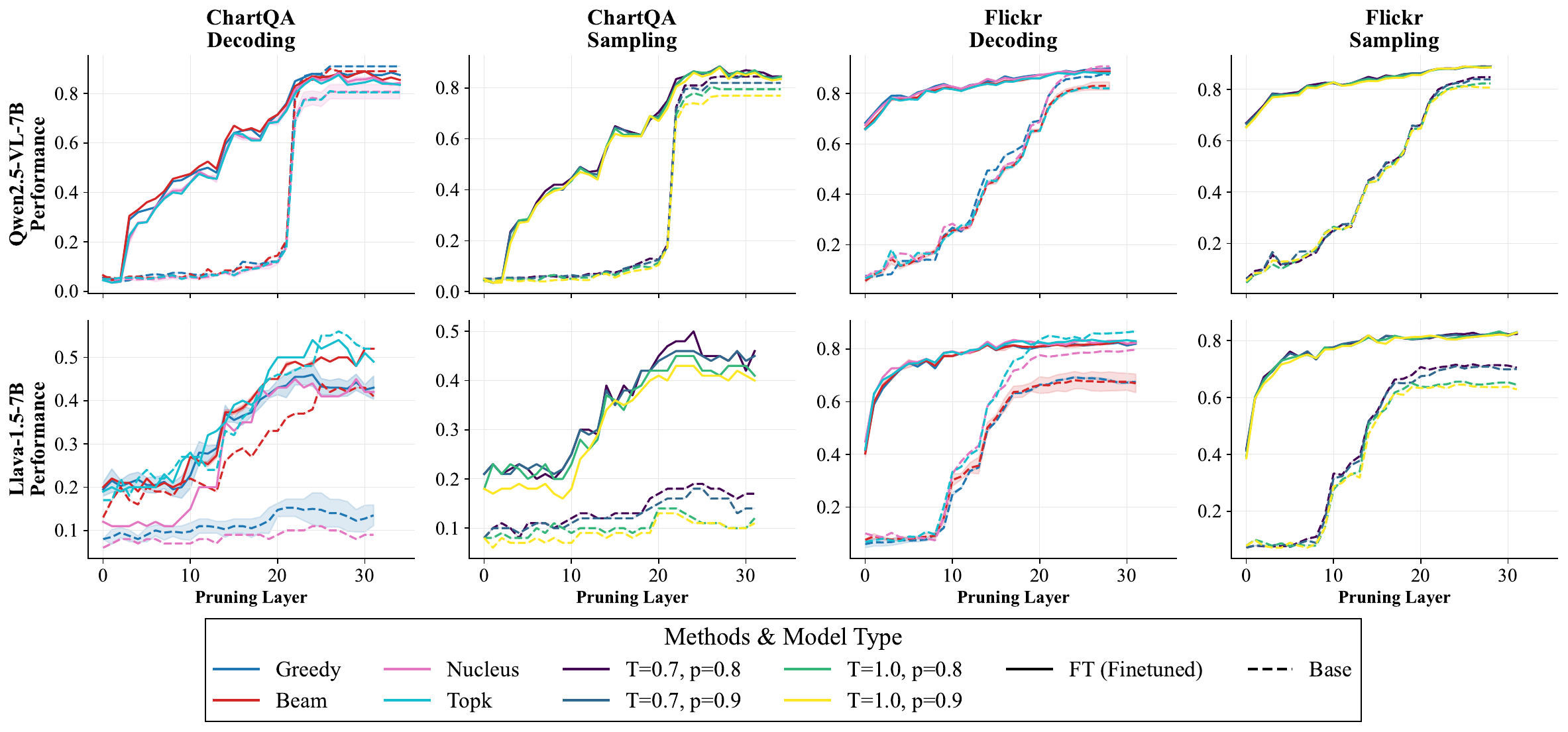}
    \caption{Performance across truncation depths under multiple decoding strategies for base and fine-tuned models. Fine-tuned models exhibit reduced dispersion across decoding configurations.}
    \label{fig:robust_perf}
\end{figure}

\begin{figure}[t]
    \centering
    \includegraphics[width=\columnwidth]{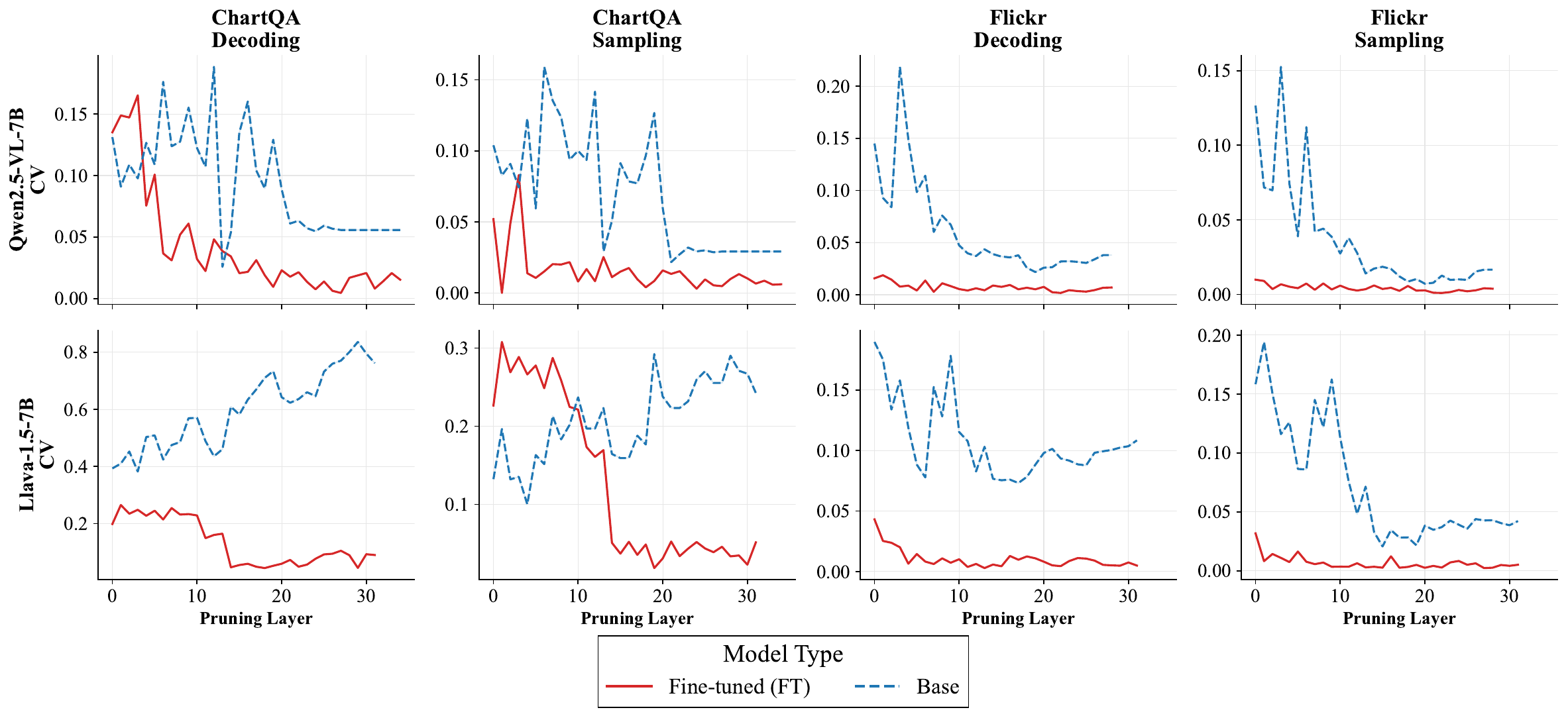}
    \caption{Coefficient of variation (CV) across decoding strategies. Fine-tuned models consistently show lower variability than truncated base models.}
    \label{fig:robust_cv}
\end{figure}

\looseness=-1\xhdr{Findings} Figs.~\ref{fig:robust_perf}-\ref{fig:robust_cv} show two consistent patterns across datasets and architectures: truncated base models exhibit high variability across decoding configurations, especially at shallow depths, whereas fine-tuned models display markedly reduced coefficient of variation across all depths. 
Fine-tuning therefore leads to improved alignment with the base model outputs and reduced cross-decoding variability under stochastic generation. This indicates that adaptation after visual-depth truncation not only restores average performance but also regularizes generation behavior, making outputs less sensitive to decoding strategy.\vspace{0.03in}

\begin{figure*}[t]
    \centering
    \includegraphics[width=\textwidth]{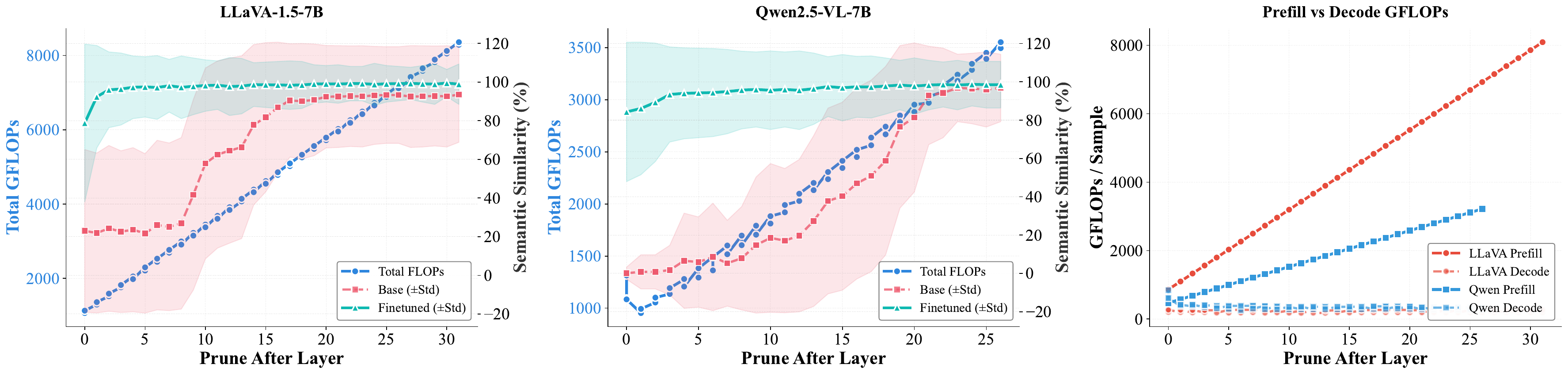}
    \caption{
    Compute–performance trade-off under visual-depth truncation.
    Blue: total GFLOPs per sample.
    Red: truncated base model performance.
    Green: fine-tuned model performance.
    Rightmost panel: prefill vs.\ decode compute decomposition.
    Fine-tuned models achieve high functional performance at substantially lower retained depth and compute budgets. Additional result on Appendix-\ref{app:rq4}
    }
    \label{fig:compute_efficiency}
\end{figure*}

\looseness=-1\noindent\textbf{\textit{Compute–performance efficiency after recovery.}} While the above results establish that fine-tuning restores functional alignment after visual-depth truncation, we now examine \textit{whether this recovery is computationally efficient}. For truncation depth $K$, let $\mathrm{GFLOPs}(K)$ denote total floating-point operations per sample and $\mathcal{M}(K)$ the task-specific performance metric (exact match for VQA; semantic similarity for captioning). Autoregressive inference is decomposed as: $\mathrm{GFLOPs}(K) = \mathrm{GFLOPs}_{\text{prefill}}(K) + \mathrm{GFLOPs}_{\text{decode}}(K),$ where pre-fill processes image and textual tokens jointly, and decode generates tokens autoregressively. Fig.~\ref{fig:compute_efficiency} reveals three consistent patterns across datasets and architectures.




\looseness=-1\xhdr{Findings} Total compute increases almost linearly with retained visual depth: $\frac{d}{dK}\mathrm{GFLOPs}(K) > 0.$ Thus, truncating visual processing yields predictable reductions in inference cost. However, models exhibit markedly different performance behavior under identical compute scaling. Truncated base models remain near-chance until deeper visual layers are restored, whereas fine-tuned models recover functional competence at substantially shallower depths. Formally, for a fixed compute budget $C$, $\mathcal{M}_{\text{FT}}(C) > \mathcal{M}_{\text{Base}}(C).$ fine-tuning therefore shifts the compute–performance frontier, achieving higher accuracy per unit of compute under truncation. Further, we observe that efficiency gains arise primarily from reductions in the prefill stage and image-token truncation significantly decreases prefill computation, while decode cost remains largely unchanged:
\[
\frac{d}{dK}\mathrm{GFLOPs}_{\text{prefill}}(K)
\gg
\frac{d}{dK}\mathrm{GFLOPs}_{\text{decode}}(K).
\]
This indicates that the computational benefits of truncation stem mainly from reduced visual grounding during prefill rather than changes in autoregressive decoding.





\begin{tcolorbox}
    \textbf{Key Takeaway.}  
Image-token processing is partially recoverable through distillation, especially for descriptive tasks, but recoverability scales with retained depth and remains incomplete under early truncation. Fine-tuning can redistribute capacity and stabilize outputs, yet sufficient visual depth is structurally necessary for precise visual grounding.
\end{tcolorbox}

\section{RQ5: Does explicit reasoning compensate for reduced visual depth?}
\label{sec:rq5_reasoning}
\looseness=-1\xhdr{Motivation} RQ3 showed multi-token generation is more sensitive to image-token truncation than single-token prediction. Next, we examine whether structured reasoning mitigates/amplifies this sensitivity by asking: \textit{Does generating explicit reasoning improve robustness to reduced visual-token processing depth?}

\xhdr{Experimental Setup} We conduct these experiments on the M3CoT~\cite{chen2024m3cot} dataset using two instruction-tuned models, LLaVA-1.5-7B~\cite{liu2024improved} and Qwen2.5-VL-7B-Instruct~\cite{bai2025qwen25vltechnicalreport}. To evaluate the effect of truncation depth $K$, we compare two generation regimes: i) single-token prediction of the final answer, and ii) structured reasoning-chain generation~\cite{xu2025llava}, which produces \texttt{<summary>}, \texttt{<caption>}, \texttt{<reasoning>} outputs prior to the final \texttt{<conclusion>}. To analyze internal degradation, we evaluate \texttt{<summary>}, \texttt{<caption>}, and \texttt{<reasoning>} using semantic similarity, BLEU, and ROUGE relative to the full base model, and compute exact match on the final \texttt{<conclusion>}  (see Appendix-\ref{app:metric}).

\begin{figure}[t]
    \centering
    \includegraphics[width=\columnwidth]{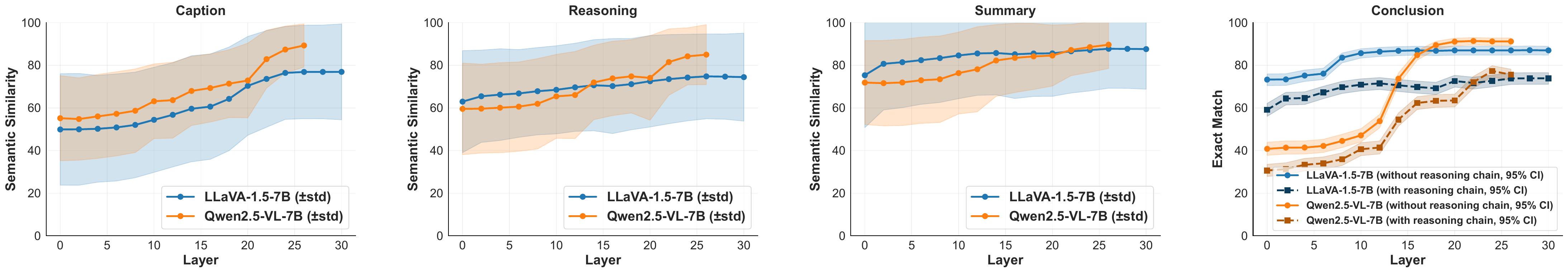}
    \caption{
    Retention of reasoning-chain components under visual-token truncation.
    Semantic similarity is computed for summary, caption,
    and reasoning relative to the base model output (see Appendix-\ref{app:rq5} for additional results).
    Caption exhibits the strongest degradation, followed by reasoning,
    while summary remains comparatively stable.
    }
    \label{fig:reasoning_component_retention}
\end{figure}

\paragraph{\textbf{Hierarchical Sensitivity Across Reasoning Components:}}
Figure \ref{fig:reasoning_component_retention} shows across truncation depths, a consistent hierarchy emerges: Caption > Reasoning > Summary. The summary remains relatively stable under early truncation, suggesting that coarse semantic abstraction is encoded early, whereas the caption degrades most strongly due to its reliance on fine-grained visual–text grounding; reasoning exhibits intermediate sensitivity, reflecting its dependence on caption-level information. Overall, visual-depth truncation induces hierarchical rather than uniform degradation across reasoning stages.

\looseness=-1 \paragraph{\textbf{Single Answer vs.\ Reasoning Chain:}} Let $A_{\text{single}}(K)$ denote Exact Match under single-token generation and $A_{\text{chain}}(K)$ denote Exact Match when the reasoning chain is generated prior to the final answer. Let $A_{\text{single}}(K)$ and $A_{\text{chain}}(K)$ denote Exact Match under single-token and reasoning-chain generation, respectively. Empirically, in Fig.~\ref{fig:reasoning_component_retention} conclusion plot, under aggressive truncation, $A_{\text{single}}(K) > A_{\text{chain}}(K)$, indicating that structured reasoning does not compensate for reduced visual depth and is in fact more sensitive to truncation. While single-token prediction requires forming a visual-conditioned decision once, reasoning-chain generation repeatedly conditions on visual information across multiple decoding steps, thereby increasing reliance on sustained visual grounding. Hence,
\[
\text{More structured decoding} \;\Rightarrow\; \text{Greater depth dependence}.
\]
Importantly, degradation in intermediate explanation quality does not always induce final answer failure, suggesting partial decoupling between representational fidelity and decision correctness.

\paragraph{Relation to Representational stabilization:}
Representation metrics characterise image-token stabilization as a representational property capturing global spectral structure and trajectory smoothness, but they do not guarantee inferential sufficiency. While stabilized representations may support single-token decisions that rely on coarse abstraction, structured reasoning requires repeated cross-modal grounding across multiple decoding steps. Consequently, even under structural stability, early visual truncation degrades multi-step reasoning, revealing residual depth dependence beyond representational consolidation.

\begin{tcolorbox}
\textbf{Key Takeaway.}
Visual representational stabilization reflects bounded structural change, not inferential completeness. Sustained visual depth remains necessary for iterative, grounded reasoning.
\end{tcolorbox}
 \section{Conclusion}
 \looseness=-1 In this work, we investigate whether large vision–language models require sustained visual-token processing across depth. We show that visual-token representations stabilize in mid-layers, where matrix entropy stabilizes, intrinsic dimensionality compresses, and trajectory curvature flattens, enabling cross-layer functional interchangeability. However, this stabilization does not imply universal redundancy. Depth truncation experiments reveal that visual-token necessity is task-dependent: single-token prediction remains relatively robust, whereas multi-token generation requires sustained visual grounding. Fine-tuning can partially restore degraded behavior after truncation, though recovery is gradual and limited for performance-sensitive tasks. Moreover, explicit reasoning does not compensate for reduced visual depth; truncation progressively degrades caption grounding and reasoning quality while summaries remain comparatively stable. Together, these findings reveal a structural decoupling between representation geometry, functional necessity, and computational allocation in multimodal transformers.


\xhdr{Why This Study Is Useful} Our study provides  the first concrete guidance for both researchers and practitioners in understanding representation dynamics of VLMs. First, it clarifies when visual depth can be safely reduced (\eg short-answer prediction) and when it must be preserved (\eg reasoning-heavy generation, less textual bias), enabling task-aware efficiency trade-offs instead of heuristic token reduction. Second, it introduces representation stabilization as a measurable signal for identifying depth regions of bounded representational change, offering a principled diagnostic tool for analyzing multimodal architectures. Third, by revealing that visual computation is predominantly concentrated in the prefill phase, it informs deployment decisions under latency or compute constraints, particularly in large-scale inference scenarios. More broadly, our results encourage evaluation of multimodal systems not only by accuracy, but by the alignment between representational dynamics, inferential demands, and computational cost, supporting the development of architectures that are structurally efficient rather than uniformly over-allocated.

\section*{Acknowledgements}
We sincerely thank Dr Suraj Srinivas for insightful discussions and valuable feedback throughout this project. 

\bibliographystyle{unsrtnat}
\bibliography{reference}
\appendix
\section{Additional Results}

\subsection{RQ1: Layer-wise Representation Dynamics}
\label{app:rq1}
\begin{figure}[h]
\centering
\includegraphics[width=\columnwidth]{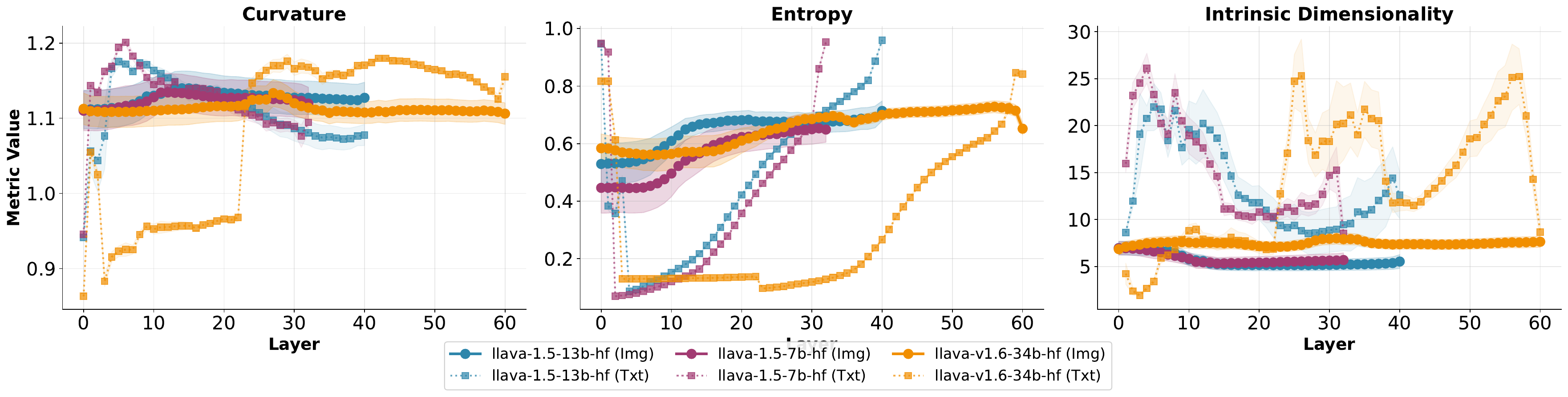}
\caption{
Layer-wise evolution of representation metrics (section~\ref{app:rq1}), for the LLaVA family.
Image tokens rapidly stabilize in early layers across all three metrics,
while textual tokens continue to evolve across depth.}
\label{fig:rq1_llava}
\end{figure}

\begin{figure}[h]
\centering
\includegraphics[width=\columnwidth]{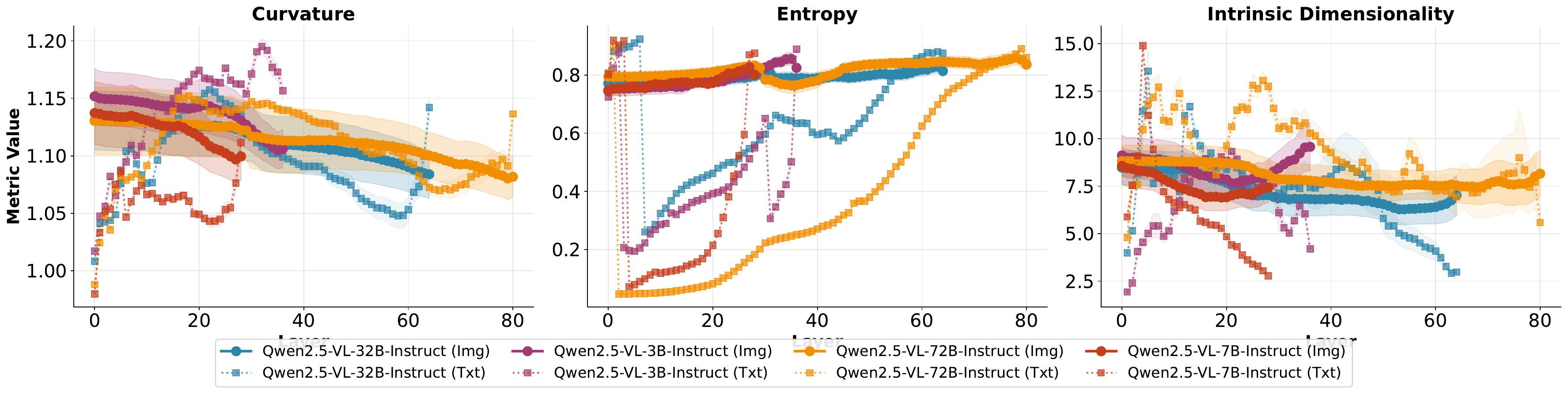}
\caption{
Representation metrics across layer-wise depth for Qwen2.5-VL models showing the
same stabilization behaviour observed in Fig.~\ref{fig:rq1_llava}.}
\label{fig:rq1_qwen25}
\end{figure}

\begin{figure}[h]
\centering
\includegraphics[width=\columnwidth]{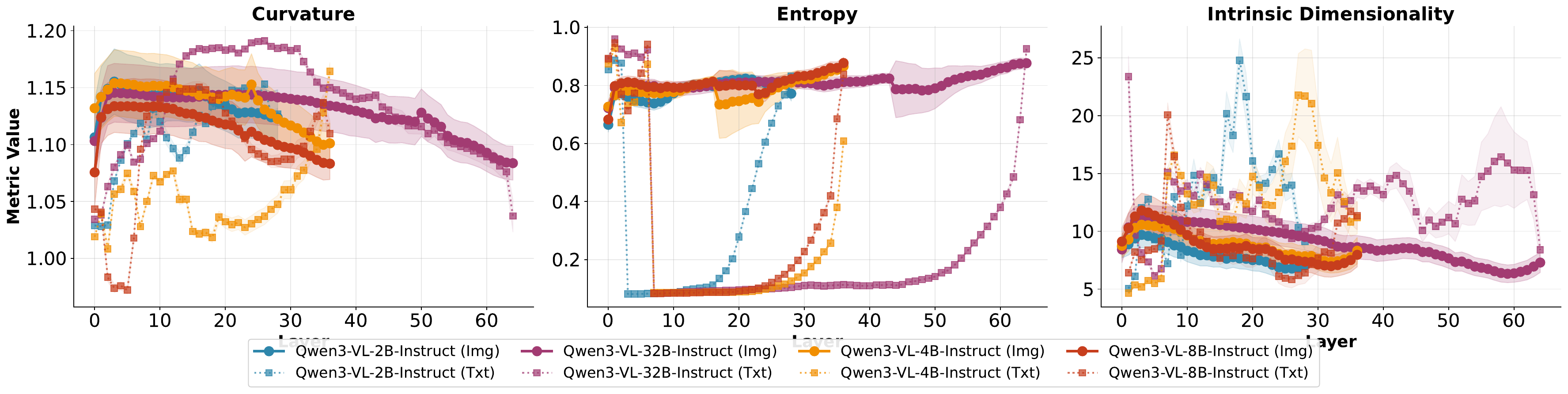}
\caption{
Layer-wise representation dynamics for the Qwen3-VL family with similar
early stabilization of visual tokens and continued evolution of textual
representations.}
\label{fig:rq1_qwen3}
\vspace{-1cm}
\end{figure}

\begin{figure}[h]
\centering
\includegraphics[width=\columnwidth]{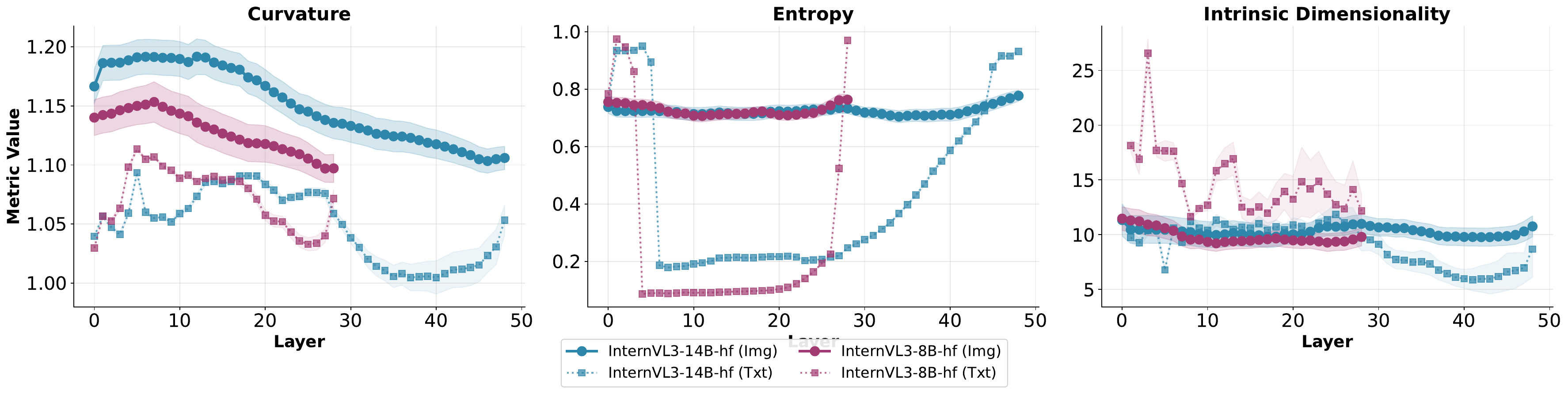}
\caption{
Representation metrics for InternVL models showing consistent
visual-token stabilization across architectures.}
\label{fig:rq1_internvl}
\end{figure}

Figures~\ref{fig:rq1_llava}, \ref{fig:rq1_qwen25}, \ref{fig:rq1_qwen3} and \ref{fig:rq1_internvl} show the layer-wise progression of representation metrics across different models.

\subsection{RQ2: Layer-wise Functional Interchangeability}
\label{app:rq2}

\begin{figure}[h]
\centering
\includegraphics[width=\columnwidth]{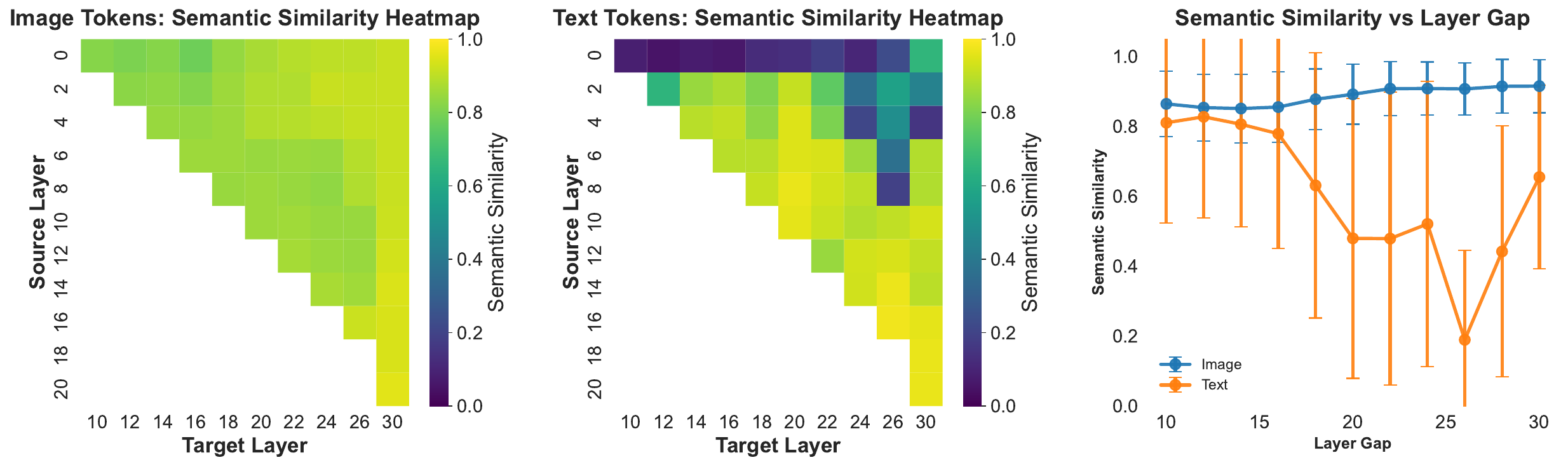}
\includegraphics[width=\columnwidth]{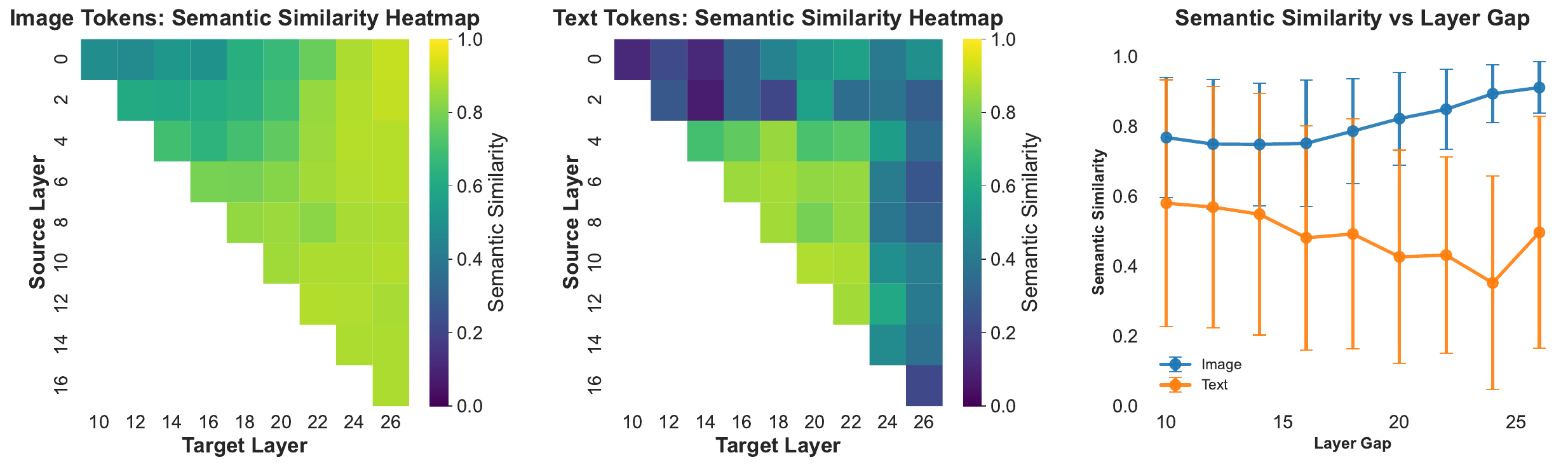}
\caption{
Layer substitution results for caption generation in LLaVA and InternVL.
Image-token substitutions preserve semantic similarity (Appendix \ref{app:metric}) across layers,
while textual substitutions lead to progressively larger degradation
as layer gaps increase.}
\label{fig:rq2_caption}
\end{figure}

To evaluate whether stabilized visual representations are functionally interchangeable across depth, we substitute representations from different layers and measure the resulting outputs using the metrics defined in Appendix~\ref{app:metric}. As shown in Figures \ref{fig:rq2_vqa} and \ref{fig:rq2_caption}, image representation substitution across layers leads to minimal changes in the output, indicating a high degree of functional interchangeability.
\begin{figure}[h]
\centering
\includegraphics[width=\columnwidth]{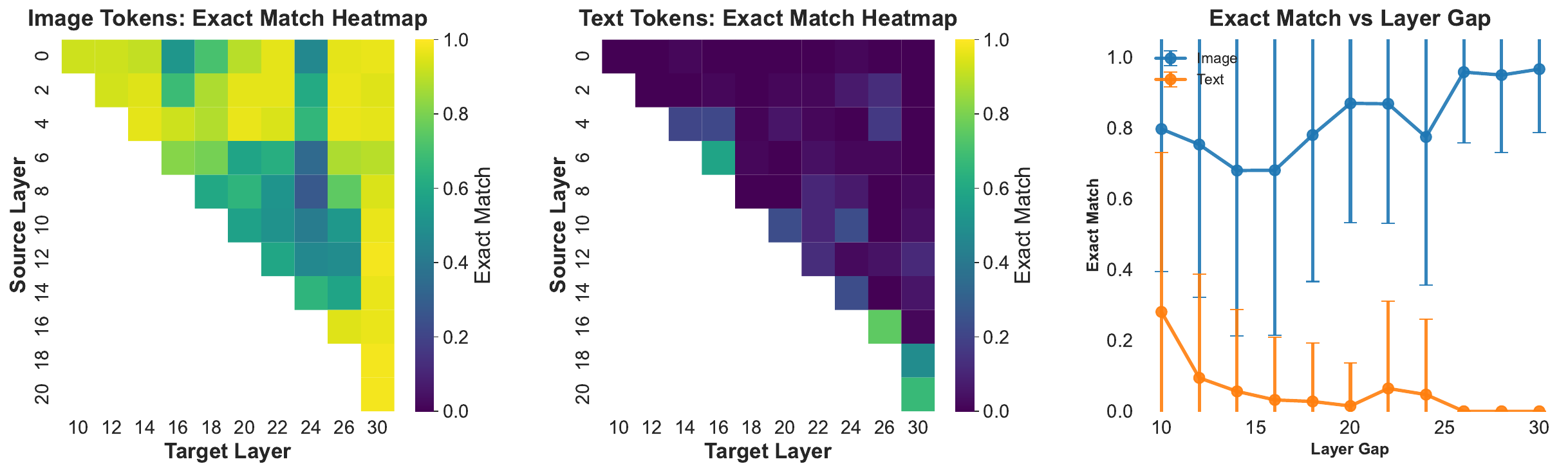}
\includegraphics[width=\columnwidth]{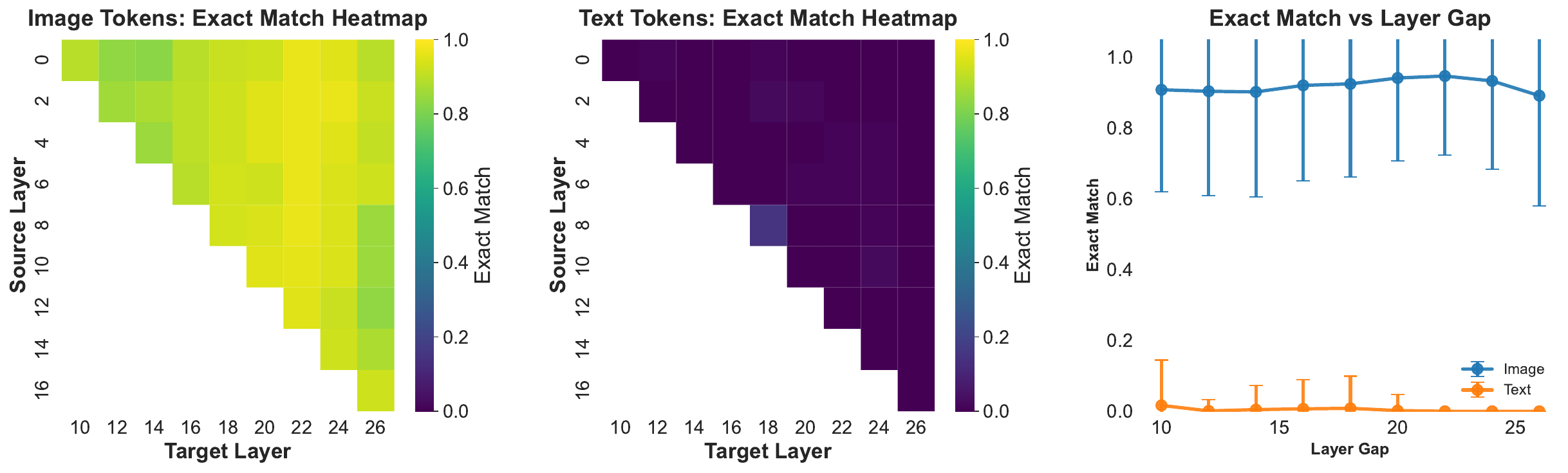}
\includegraphics[width=\columnwidth]{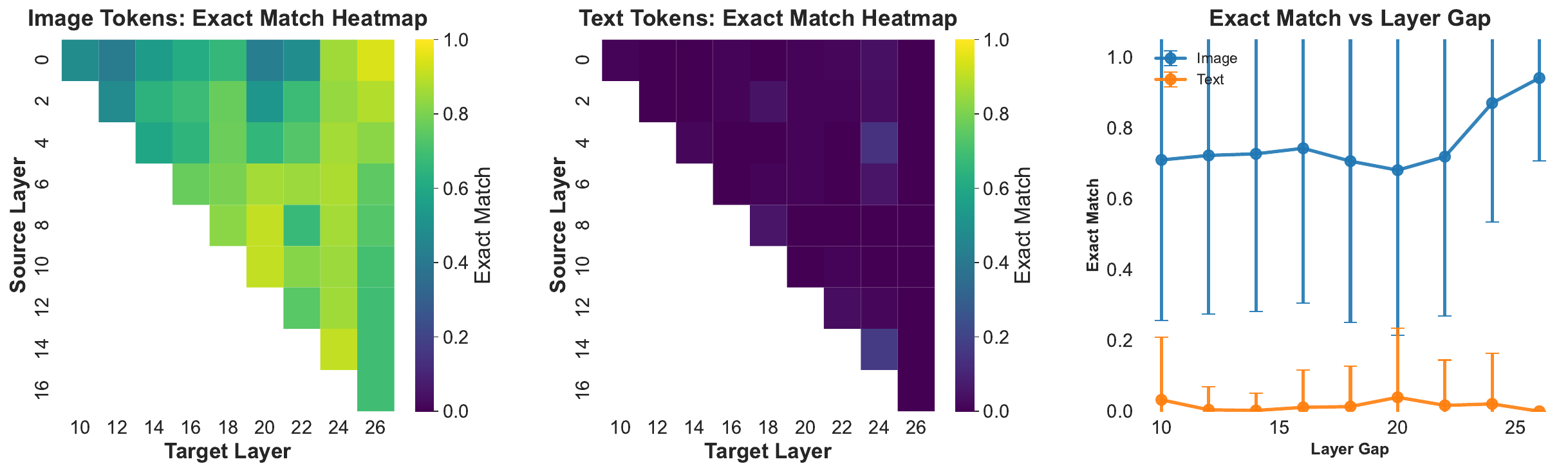}
\caption{
Cross-layer substitution experiments for VQA-style tasks for LLaVA, Qwen2.5-VL and InternVL.
Replacing image tokens with earlier-layer representations causes minimal
degradation in exact match (Appendix \ref{app:metric}) accuracy, whereas substituting textual tokens
results in larger performance drops.}
\label{fig:rq2_vqa}
\vspace{-1cm}
\end{figure}

\subsection{RQ3: Task-Dependent Utility of Image Tokens}
\label{app:rq3}

To examine how the necessity of visual processing varies across tasks, we evaluate model performance under progressive truncation of image-token processing depth. Performance is measured using exact match for bounded generation tasks and semantic similarity for open-ended generation, as defined in Appendix~\ref{app:metric}.



\subsection{RQ4: Recovery Through Fine-Tuning}
\label{app:rq4}
\begin{figure}[h]
\centering
\includegraphics[width=\columnwidth]{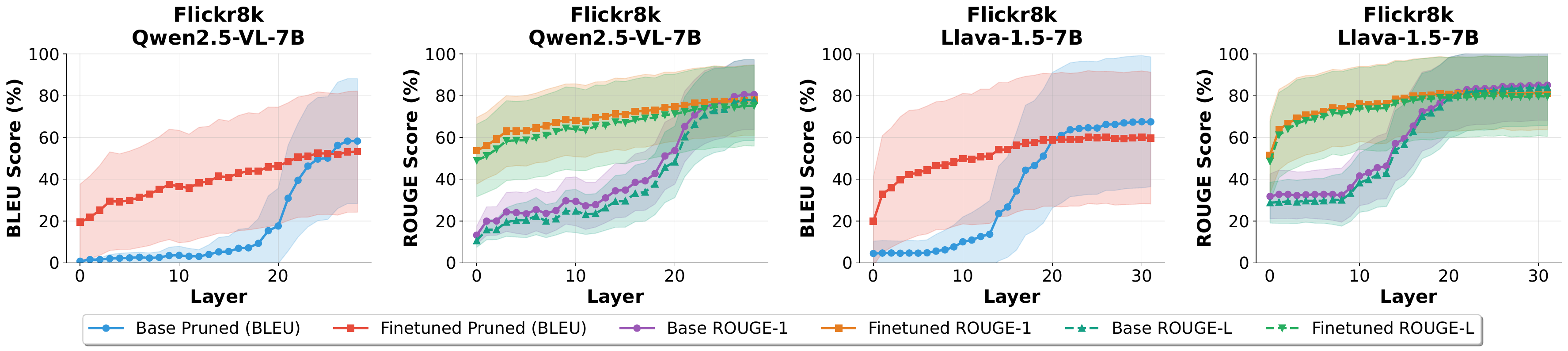}
\caption{
Caption generation performance using BLUE and ROGUE scores on Flickr8K under visual processing depth truncation. Fine-tuned models significantly improve semantic similarity with the
base model outputs compared to truncated models.}
\label{fig:rq4_flickr}
\end{figure}

\begin{figure}
\centering
\includegraphics[width=\columnwidth]{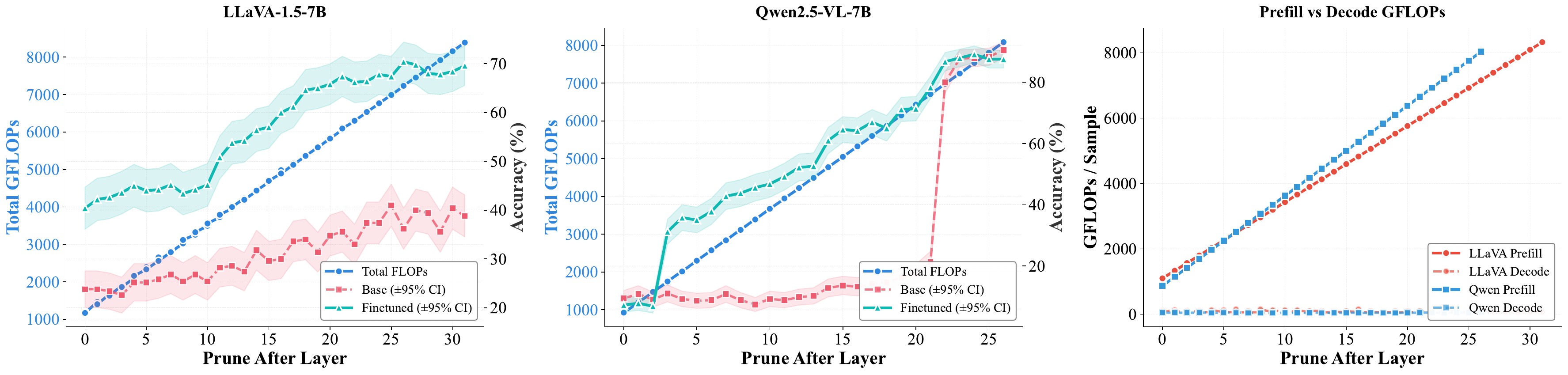}
\caption{
ChartQA VQA performance vs compute flops using exact match (Appendix \ref{app:metric}) under image-token truncation.
Exact match accuracy improves after fine-tuning with fewer flops. Also, prefill phase benefits more compare to decode phase. }
\label{fig:rq4_chartqa}
\end{figure}

We investigate whether truncated models can recover the behavior of the original model through distillation-based fine-tuning. We evaluate this by measuring the lexical and semantic similarity between outputs of the fine-tuned truncated model and the base model. Figure~\ref{fig:rq4_flickr} reports BLEU and ROUGE scores for captioning, while Figure~\ref{fig:rq4_chartqa} shows the compute advantages for VQA after fine-tuning. VQA performance is measured using exact match, and captioning using semantic similarity, as defined in Appendix~\ref{app:metric}.

\subsection{RQ5: Reasoning Sensitivity to Visual Depth}
\label{app:rq5}

\begin{figure}
\centering
\includegraphics[width=\columnwidth]{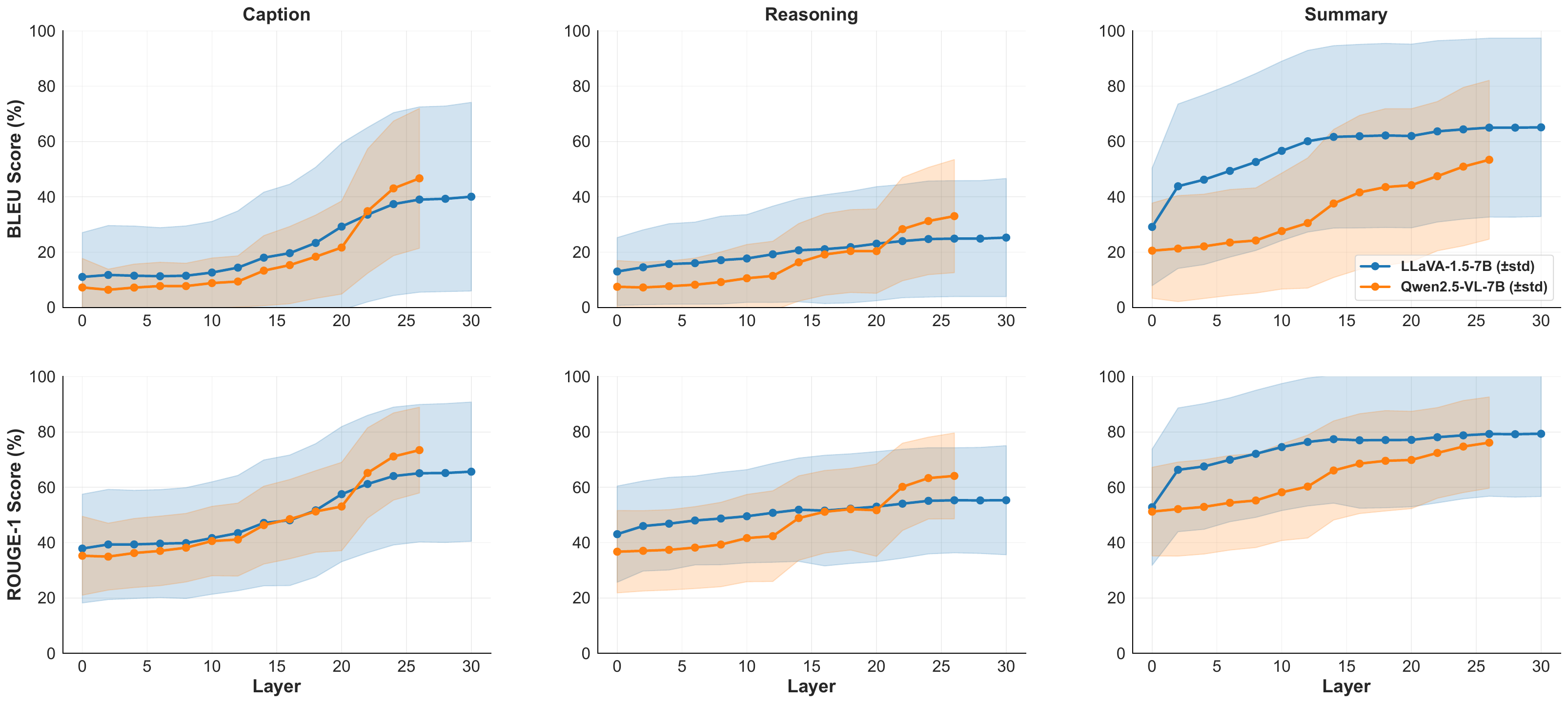}
\caption{
BLEU and ROGUE similarity of reasoning-chain components under visual-token
truncation. The caption shows the strongest degradation, followed by
reasoning, while the summary remains comparatively stable.}
\label{fig:rq5_reasoning}
\end{figure}

We further analyze how reasoning-chain generation behaves under reduced visual-token processing depth. Figure~\ref{fig:rq5_reasoning} illustrates the layerwise lexical progression across different stages of the reasoning process.

\subsection{Evaluation Metrics}
\label{app:metric}

We evaluate model behavior using two complementary metrics.
\begin{enumerate}
    \item \xhdr{Exact Match} Exact Match measures whether the predicted answer $\hat{y}_i$ exactly matches the reference answer $y_i$. Over a dataset of $N$ samples, the Exact Match score is defined as $\frac{1}{N} \sum_{i=1}^{N} \mathbf{1}(\hat{y}_i = y_i)$, where $\mathbf{1}(\cdot)$ is the indicator function that returns
$1$ if the predicted answer exactly matches the reference answer
and $0$ otherwise. "Accuracy" and "exact match" have been used interchangeably.

\item \xhdr{Semantic Similarity} For open-ended generation tasks we measure semantic similarity between the generated output $\hat{s}_i$ and the reference output $s_i$ using cosine similarity between sentence embeddings. Embeddings are obtained using a Sentence-Transformer model. The similarity is defined as:

\begin{equation}
    \text{SS}(\hat{s}_i, s_i) = \frac{f(\hat{s}_i) \cdot f(s_i)} {\|f(\hat{s}_i)\| \, \|f(s_i)\|},
\end{equation}
where $f(\cdot)$ denotes the embedding function  by the Sentence-Transformer model.
\end{enumerate}



\end{document}